\documentclass{bmvc2k}

\title{Rethinking Curriculum Learning with Incremental Labels and Adaptive Compensation}

\addauthor{Madan Ravi Ganesh}{madantrg@umich.edu}{1}
\addauthor{Jason J. Corso}{jjcorso@umich.edu}{1}

\addinstitution{
University of Michigan\\
EECS\\
Ann Arbor\\
Michigan, USA}

\runninghead{Ravi Ganesh, Corso}{Label-based Curriculum Learning}

\newcommand{\alg}{LILAC}
\DeclareMathOperator*{\argmaxA}{arg\,max} 

\usepackage{graphicx}
\usepackage{amsmath,amssymb} % define this before the line numbering.
\usepackage{multirow}
\usepackage{color, colortbl}
\definecolor{Gray}{gray}{0.9}
\usepackage{enumitem}
\usepackage{booktabs}
\usepackage{bbm}
\usepackage{pifont}
\usepackage{bbold}
\usepackage{caption}
\def\eg{\emph{e.g}\bmvaOneDot}
\def\Eg{\emph{E.g}\bmvaOneDot}
\def\etal{\emph{et al}\bmvaOneDot}

\begin{document}

\maketitle

\begin{abstract}
Like humans, deep networks have been shown to learn better when samples are organized and introduced in a meaningful order or curriculum~\cite{weinshall2018curriculum}.
Conventional curriculum learning schemes introduce samples in their order of difficulty.
This forces models to begin learning from a subset of the available data while adding the external overhead of evaluating the difficulty of samples.
In this work, we propose \textit{Learning with Incremental Labels and Adaptive Compensation} (\alg), a two-phase method that incrementally increases the number of unique output labels rather than the difficulty of samples while consistently using the entire dataset throughout training.
In the first phase, \textit{Incremental Label Introduction}, we partition data into mutually exclusive subsets, one that contains a subset of the ground-truth labels and another that contains the remaining data attached to a pseudo-label.
Throughout the training process, we recursively reveal unseen ground-truth labels in fixed increments until all the labels are known to the model.
In the second phase, \textit{Adaptive Compensation}, we optimize the loss function using altered target vectors of previously misclassified samples. 
The target vectors of such samples are modified to a smoother distribution to help models learn better.
On evaluating across three standard image benchmarks, CIFAR-10, CIFAR-100, and STL-10, we show that \alg{} outperforms all comparable baselines.
Further, we detail the importance of pacing the introduction of new labels to a model as well as the impact of using a smooth target vector.
\end{abstract}

%-------------------------------------------------------------------------
\section{Introduction}
\label{sec:introduction}
% Setup of problem
Deep networks are a notoriously hard class of models to train effectively~\cite{erhan2009difficulty,larochelle2007empirical,glorot2010understanding,larochelle2009exploring}.
A combination of high-dimensional problems, characterized by a large number of labels and a high volume of samples, a large number of free parameters and extreme sensitivity to experimental setups are some of the main reasons for the difficulty in training deep networks.
The go-to solution for deep network optimization is Stochastic Gradient Descent with mini-batches~\cite{robbins1951stochastic} (batch learning) or its derivatives.
There are two alternative lines of work which offer strategies to guide deep networks to better solutions than batch learning: Curriculum Learning~\cite{bengio2009curriculum,florensa2017reverse,graves2017automated} and Label Smoothing~\cite{elman1993learning,xie2016disturblabel}.

% Current approaches
Curriculum learning helps deep networks learn better by gradually increasing the difficulty of samples used to train networks.
This idea is inspired by methods used to teach humans and patterns in human cognition and behaviour~\cite{skinner1958reinforcement,avrahami1997teaching}.
The ``difficulty'' of samples in the dataset, obtained using either external ranking methods or internal rewards~\cite{hacohen2019power,florensa2017reverse}, introduces an extra computational overhead while the setup itself restricts the amount of data from which the model begins to learn.

Label smoothing techniques~\cite{xie2016disturblabel,reed2014training,pereyra2017regularizing} regularize the outcomes of deep networks to prevent over-fitting while improving on existing solutions.
They penalize network outputs based on criteria such as noisy labels, overconfident model outcomes, or robustness of a network around a data point in the feature space.
Often, such methods penalize the entire dataset throughout the training phase with no regard to the prediction accuracy of each sample.

\begin{figure}[t]
    \begin{center}
    \includegraphics[width=\columnwidth]{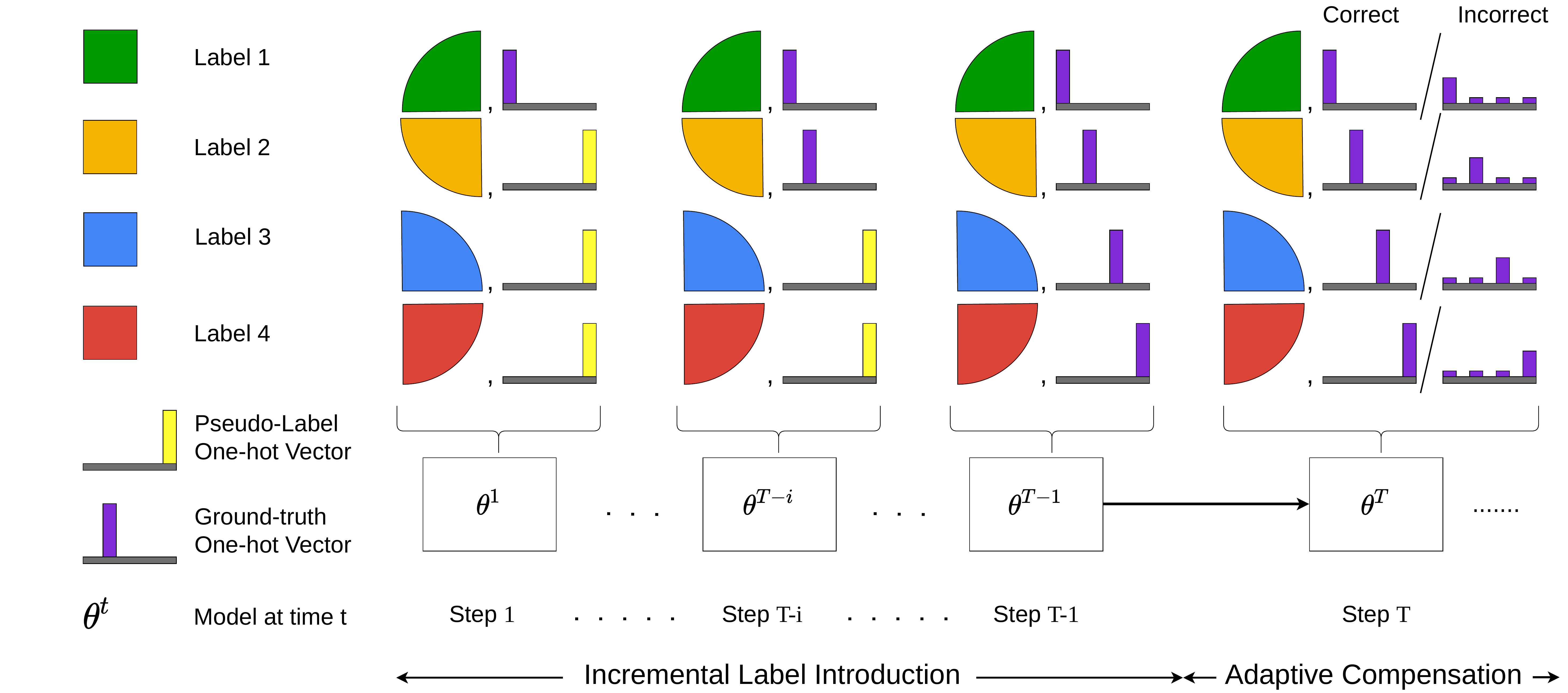}
    \end{center}
    \caption{Illustration of the components of \alg for a four label dataset case. The \textit{Incremental Label introduction} (IL) phase introduces new labels at regular intervals while using the data corresponding to unknown labels (pseudo-label) as negative samples. Once all the labels have been introduced, the \textit{Adaptive Compensation} (AC) phase of training begins. Here, a prior copy of the network is used to classify training data. If a sample is misclassified then a smoother distribution is used as its ground-truth vector in the current epoch.}
    \label{fig:lilac_concept}
\end{figure}

% Our approach
Inspired by an alternative outlook on Elman's~\cite{elman1993learning} notion of ``starting small'', we propose \alg, \textit{Learning with Incremental Labels and Adaptive Compensation}, a novel label-based algorithm that overcomes the issues of the previous methods and effectively combines them.
\alg{} works in two phases, 1) \textit{Incremental Label Introduction} (IL), which emphasizes gradually learning labels, instead of samples, and 2) \textit{Adaptive Compensation} (AC), which regularizes the outcomes of previously misclassified samples by modifying their target vectors to smoother distributions in the objective function (Fig.~\ref{fig:lilac_concept}).

In the first phase, we partition data into two mutually exclusive sets: $\mathbb{S}$, a subset of ground-truth (GT) labels and their corresponding data; and $\mathbb{U}$, remaining data associated with a pseudo-label ($\rho$) and used as negative samples.
Once the network is trained using the current state of the data partition for a fixed interval, we reveal more GT labels and their corresponding data and repeat the training process.
By contrasting data in $\mathbb{S}$ against the entire remaining dataset in $\mathbb{U}$, we consistently use all the available data throughout training, thereby overcoming one of the key issues of curriculum learning.
The setup of the IL phase, inspired by continual learning, allows us to flexibly space out the introduction of new labels and provide the network enough time to develop a strong understanding of each class.

Once all the GT labels are revealed, we initiate the AC phase of training.
In this phase, we replace the target one-hot vector of misclassified samples, obtained from a previous version of the network being trained,  with a smoother distribution.
The smoother distribution provides an easier value for the network to learn while the use of a prior copy of the network helps avoid external computational overhead and limits the alteration to only necessary samples.

% Contributions/Claims highlights
To summarize, our main contributions in \alg{} are as follows:
\begin{itemize}[topsep=0pt,itemsep=-1ex,partopsep=1ex,parsep=1ex]
    \item we introduce a novel method for curriculum learning that incrementally learns \textit{labels} as opposed to \textit{samples},
    \item we formulate \textit{Adaptive Compensation} as a method to regularize misclassified samples while removing external computational overhead,
    \item finally, we improve average recognition accuracy across all of the evaluated benchmarks compared to batch learning, a property that is not shared by the other tested curriculum learning and label smoothing methods.
\end{itemize}

\noindent Our code is available at \url{https://github.com/MichiganCOG/LILAC_v2}.

\section{Related Works}
\label{sec:related_works}
\paragraph{Curriculum Learning}
Bengio \etal{~\cite{bengio2009curriculum}}, Florensa \etal{~\cite{florensa2017reverse}}, and Graves \etal{~\cite{graves2017automated}} are some important works that have redefined and applied curriculum learning in the context of deep networks.
These ideas were expanded upon to show improvements in performance across corrupted~\cite{jiang2017mentornet} and small datasets~\cite{fan2018learning}.
More recently, Hacohen and Weinshall~\cite{hacohen2019power} explored the impact of varying the pace with which samples were introduced while Weinshall~\cite{weinshall2018curriculum} used alternative deep networks to categorize difficult samples.
To the best of our knowledge, most previous works have assumed that samples cover a broad spectrum of difficulty and hence need to be categorized and presented in an orderly fashion.
The closest relevant work to ours, in terms of learning labels, gradually varies the GT vector from a multimodal distribution to a one-hot vector over the course of the training phase~\cite{dogan2019label}.

\paragraph{Label Smoothing}
Label smoothing techniques regularize deep networks by penalizing the objective function based on a pre-defined criterion. 
Such criteria include using a mixture of true and noisy labels~\cite{xie2016disturblabel}, penalizing highly confident outputs~\cite{pereyra2017regularizing}, and using an alternate deep network's outcomes as GT~\cite{reed2014training}.
Bagherinezhad \etal{~\cite{bagherinezhad2018label}} proposed the idea of using logits from trained models instead of just one-hot vectors as GT.
Complementary work by Miyato \etal{~\cite{miyato2015distributional}} used the local distributional smoothness, based on the robustness of a model's distribution around a data point, to smooth labels.
The work closest to our method was proposed in Szegedy \etal{~\cite{szegedy2016rethinking}}, where an alternative target distribution was used across the entire dataset.
Instead, we propose to alter the GT vector for only samples that are misclassified. 
They are identified using a prior copy of the current model, which helps avoid external computational overhead and only uses a small set of operations.

\paragraph{Incremental Learning and Negative Mining}
Incremental and Continual learning are closely related fields that inspired the structure of our algorithm.
Their primary concern is learning over evolving data distributions with the addition of constraints on the storage memory~\cite{rebuffi2017icarl, castro2018end}, distillation of knowledge across different distributions~\cite{schwarz2018progress,rolnick2019experience}, assumption of a single pass over data~\cite{lopez2017gradient,chaudhry2018efficient}, etc.
In our approach, we depart from the assumption of evolving data distributions.
Instead, we adopt the experimental pipeline used in incremental learning to introduce new labels at regular intervals.
At the same time, inspired by negative mining~\cite{10.1007/978-3-319-49409-8_45,li2013bootstrapping,wang2015unsupervised}, we use the remaining training data, associated with a pseudo-label, as negative samples.
Overall, our setup effectively uses the entire training dataset, thus maintaining the same data distribution.

\section{\alg}
\label{sec:algorithm}

In \alg, our main objective is to improve upon batch learning.
We do so by first gradually learning labels, in fixed increments, until all GT labels  are known to the network (Section~\ref{subsec:il}).
This behaviour assumes that all samples are of equal difficulty and are available to the network throughout the training phase.
Further, we focus on learning strong representations of each class over a dedicated period of time.
Once all GT labels are known, we shift to regularizing previously misclassified samples by smoothing the distribution of their target vector while maintaining the peak at the same GT label (Section~\ref{subsec:ac}).
Using a smoother distribution leads to an increase in the entropy of the target vector and helps the network learn better, as we demonstrate in Section~\ref{sec:key_properties_of_lilac}.
% An overview of the entire \alg{} algorithm is available in the supplementary materials.

\begin{figure}[t!]
    \begin{center}
    \includegraphics[width=0.9\columnwidth]{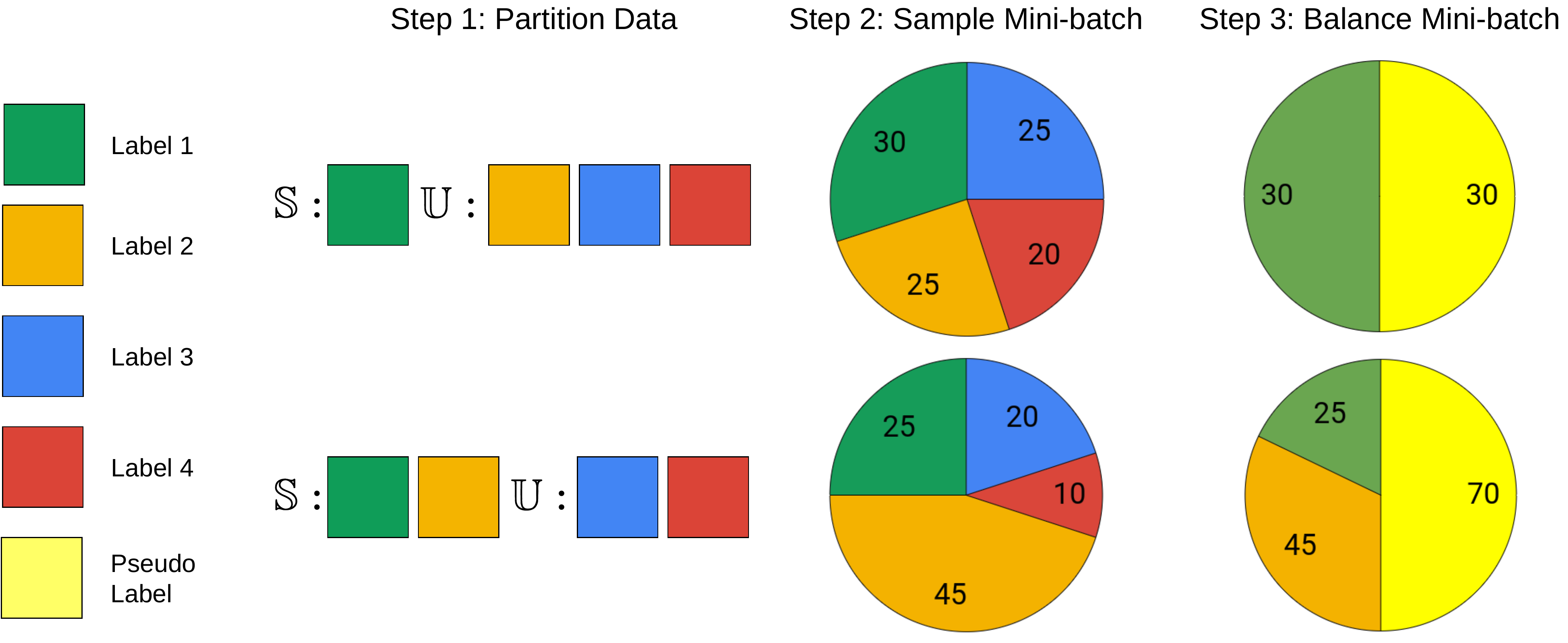}
    \end{center}
        \caption{Illustration of the steps in the IL phase when (\textbf{Top}) only one GT label is in $\mathbb{S}$ and (\textbf{Bottom}) when two GT labels are in $\mathbb{S}$.
        The steps are 1) partition data, 2) sample a mini-batch of data and 3) balance the number of samples from $\mathbb{U}$ to match those from $\mathbb{S}$ in the mini-batch before training. 
        Samples from $\mathbb{U}$ are assumed to have a uniform prior when being augmented/reduced to match the total number of samples from $\mathbb{S}$. Values inside each pie represent the number of samples. 
        Across both cases, the number of samples from $\mathbb{S}$ determines the final balanced mini-batch size.}
    \label{fig:il_phase}
\end{figure}

\subsection{Incremental Label Introduction Phase}
\label{subsec:il}
In the IL phase, we partition data into two sets: $\mathbb{S}$, a subset of GT labels and their corresponding data; and $\mathbb{U}$, the remaining data marked as negative samples using a pseudo-label $\rho$.
Over the course of multiple intervals of training, we reveal more GT labels to the network according to a predetermined schedule. 
Within a given interval of training, the data partition is held fixed and we uniformly sample mini-batches from the entire training set based on their GT label.
However, for samples from $\mathbb{U}$, we use $\rho$ as their label.
There is no additional change required in the objective function or the outputs of the model when we sample data from $\mathbb{U}$.
By the end of this phase, we reveal all GT labels to the network.

% We now describe the IL phase in detail.
For a given dataset, we assume a total of $L$ labels are provided in the ascending order of their value.
Based on this ordering, we initialize the first $b$ labels, and their corresponding data, as $\mathbb{S}$, and the data corresponding to the remaining $L-b$ labels as $\mathbb{U}$.
Over the course of multiple training intervals, we reveal GT labels in increments of $m$, a hyper-parameter that controls the schedule of new label introduction.
Revealing a GT label involves moving the corresponding data from $\mathbb{U}$ to $\mathbb{S}$ and using their GT label instead of $\rho$.

Within a training interval, we train the network for $E$ epochs using the current state of the data partition.
First, we sample a mini-batch of data based on a uniform prior over their GT labels.
Then, we modify their target vectors based on the partition to which a sample belongs.
% For a mini-batch of uniformly sampled data, based on GT labels, we modify their target vector based on the partition to which a sample belongs.
To ensure the balanced occurrence of samples from GT labels and $\rho$, we augment or reduce the number of samples from $\mathbb{U}$ to match those from $\mathbb{S}$ and use this curated mini-batch to train the network.
After $E$ epochs, we move $m$ new GT labels and their corresponding data from $\mathbb{U}$ to $\mathbb{S}$ and repeat the entire process (Fig.~\ref{fig:il_phase}).

\subsection{Adaptive Compensation}
\label{subsec:ac}
Once all the GT labels have been revealed and the network has trained sufficiently, we begin the AC phase.
In the AC phase, we use a smoother distribution for the target vector of samples which the network is unable to correctly classify.
% The main idea behind AC is, if the network is unable to correctly predict a sample's label then we should increase the entropy of its target vector to obtain a smoother distribution than one-hot.
Compared to one-hot vectors, optimizing over this smoother distribution, with an increased entropy, can bridge the gap between the unequal distances in the embedding space and overlaps in the label space~\cite{rodriguez2018beyond}.
This overlap can occur due to common image content or close proximity in the embedding space relative to other classes. 
Thus, improving the entropy of such target vectors can help modify the embedding space in the next epoch and compensate for the predictions of misclassified samples.
% Unlike prior methods, we adaptively modify the target vector, only for misclassified samples, on-the-fly. 

For a sample $(x_{i},y_{i})$ in epoch $e \geq T$, we use predictions from the model at $e-1$ to determine the final target vector used in the objective function; specifically, we smoothen the target vector for a sample if and only if it was misclassified by the model at epoch $e-1$.
Here, $(x_{i},y_{i})$ denotes a training sample and its corresponding GT label for sample index $i$, and $T$ represents a threshold epoch value until which the network is trained without adaptive compensation.
We compute the final target vector for the $i^{th}$ instance at epoch $e$, $t_{i}^{e}$, based on the model $\theta^{e-1}$ using the following equation,

\begin{equation}
    t_{i}^{e} = \begin{cases} 
            (\frac{\epsilon L - 1}{L-1}) \delta_{y_{i}} + (\frac{1 - \epsilon}{L-1}) \mathbb{1}, \quad \argmaxA\big(f_{\theta^{e-1}}(x_{i})\big) \neq y_{i} \\
            \delta_{y_{i}}, \qquad \qquad  \qquad \quad ~~~\text{otherwise}
            \end{cases}.
            \label{eq:ac}
\end{equation}

Here, $\delta_{y_{i}}$ represents the one-hot vector corresponding to GT label $y_i$, $\mathbb{1}$ is a vector of $L$ dimensions with all entries as 1 and $\epsilon$ is a scaling hyper-parameter.

\section{Experiments}
\label{sec:experiments}

\textbf{Datasets and Metrics}
We use three datasets, CIFAR-10, CIFAR-100~(\cite{krizhevsky2009learning}), and STL-10~(\cite{coates2011analysis}), to evaluate our method and validate our claims.
CIFAR-10 and CIFAR-100 are 10 and 100 class variants of the popular image benchmark CIFAR while STL-10 is a 10 class subset of ImageNet.
\textit{Average Recognition Accuracy~(\%)} combined with their \textit{Standard Deviation} across 5 trials are used to evaluate the performance of all the algorithms. 
\hfill \\

\noindent\textbf{Experimental Setup}
For CIFAR-10/100, we use ResNet18~(\cite{he2016deep}) as the architectural backbone while for STL-10, we use ResNet34.
We set $\rho$ as the last label and $b$ as half the total number of labels of a given dataset.
In each interval of \alg's IL phase, we train the model for 7, 3, and 10 epochs each, at a learning rate of 0.1, 0.01, and 0.1 for CIFAR-10, CIFAR-100, and STL-10, respectively.
In the AC phase epochs 150, 220, and 370 are used as thresholds (epoch $T$) for CIFAR-10, CIFAR-100, and STL-10 respectively.
Detailed explanations of the experimental setups are provided in the supplementary materials. 
\hfill \\

\noindent\textbf{Baselines} 
\begin{enumerate}[topsep=0pt,itemsep=-1ex,partopsep=1ex,parsep=1ex,leftmargin=*]
\item Stochastic Gradient Descent with mini-batches (Batch Learning).
\item Standard Baselines
\begin{itemize}[topsep=-2ex,itemsep=-1ex,partopsep=1ex,parsep=1ex,leftmargin=*]
    \item Fixed Curriculum: Following the methodology proposed in Bengio \etal{~\cite{bengio2009curriculum}}, we create a ``Simple'' subset of the dataset using data that is within a value of 1.1 as predicted by a linear one-vs-all SVR model.
    The deep network is trained on the ``Simple'' dataset for a fixed period of time, which mirrors the total length of the IL phase, after which the entire dataset is used to train the network. 
    \item Label Smoothing: We follow the method proposed in Szegedy \etal{~\cite{szegedy2016rethinking}}.
\end{itemize}
\item Custom Baselines
\begin{itemize}[topsep=-2ex,itemsep=-1ex,partopsep=1ex,parsep=1ex,leftmargin=*]
    \item Dynamic Batch Size (DBS): DBS randomly copies data available within a mini-batch to mimic variable batch sizes, similar to the IL phase. However, all GT labels are available to the model throughout the training process.
    \item Random Augmentation (RA): This baseline samples from a single randomly chosen class in $\mathbb{U}$, available in the current mini-batch, to balance data between $\mathbb{S}$ and $\mathbb{U}$ in the current mini-batch.
    This is in contrast to \alg{}, which uses samples from all classes in $\mathbb{U}$ that are available in the current mini-batch.
\end{itemize}
\item Ablative Baselines 
\begin{itemize}[topsep=-2ex,itemsep=-1ex,partopsep=1ex,parsep=1ex,leftmargin=*]
    \item \textit{Only IL}: This baseline quantifies the contribution of incrementally learning labels when combined with batch learning.
    \item \textit{Only AC}: This baseline shows the impact of adaptive compensation, as a label smoothing technique, when combined with batch learning. 
\end{itemize}
\end{enumerate}

\subsection{Comparison Against Standard Baselines}
\label{sec:comparison_against_standard_baselines}

Table~\ref{tab:lilac_baselines} illustrates the improvement offered by \alg{} over Batch Learning, with comparable setups.
Further, we break down the contributions of each phase of \alg.
Both \textit{Only IL} and \textit{Only AC} improve over batch learning, albeit to varying degrees, which highlights their individual strengths and importance.
However, only when we combine both phases do we observe a consistently high performance across all benchmarks.
This indicates that these two phases complement each other.

The Fixed Curriculum approach does not offer consistent improvements over the Batch Learning baseline across CIFAR-100 and STL-10 while the Label Smoothing approach does not outperform batch learning on the STL-10 dataset.
While both of these standard baselines fall short, \alg{} consistently outperforms Batch Learning across all evaluated benchmarks.
Interestingly, Label Smoothing provides the highest performance on CIFAR-100.
Since the original formulation of \alg{} was based on Batch Learning, we assumed all GT vectors to be one-hot.
This assumption is violated in Label Smoothing. 
When we tailor our GT vectors according to the Label Smoothing baseline, we outperform it with minimal hyper-parameter changes, a testament to \alg's applicability on top of conventional label smoothing.

\begin{table}[t!]
\begin{center}
\begin{tabular}{@{}llccc@{}}
\toprule
\multirow{2}{*}{Types} & \multirow{2}{*}{Training}          & \multicolumn{3}{c}{Performance (\%)}  \\ 
        &  & CIFAR 10               & CIFAR 100           & STL 10                            \\ \midrule
& Batch Learning           & 95.19 $\pm$ 0.190      &  78.32 $\pm$ 0.175   &   72.88 $\pm$ 0.642  \\ \midrule
\multirow{2}{*}{Standard} & Fixed Curriculum~\cite{bengio2009curriculum}  & 95.27 $\pm$ 0.112      &  77.89 $\pm$ 0.287   &   72.18 $\pm$ 0.601  \\
& Label Smoothing~\cite{szegedy2016rethinking} & 95.27 $\pm$ 0.111      &  79.06 $\pm$ 0.179   &   72.55 $\pm$ 0.877  \\
\multirow{2}{*}{Custom} & Random Augmentation      &  95.27 $\pm$ 0.076     &  75.37 $\pm$ 0.480   &   73.67 $\pm$ 0.708  \\
& Dynamic Batch Size       &  95.22 $\pm$ 0.131     &  78.73 $\pm$ 0.264   &   72.66 $\pm$ 1.081  \\ \midrule
\multirow{2}{*}{Ablative} & \textit{Only IL}  (\textbf{ours}) &  95.38 $\pm$ 0.135 &  78.73 $\pm$ 0.139       &   73.43 $\pm$ 0.903  \\ 
& \textit{Only AC}  (\textbf{ours}) &  95.38 $\pm$ 0.170 &  78.94 $\pm$ 0.179       &   72.94 $\pm$ 0.530  \\ 
\multirow{2}{*}{Overall} & \alg{} (\textbf{ours})     & \textbf{95.52 $\pm$ 0.072}  &  78.88 $\pm$ 0.201       & \textbf{73.77 $\pm$ 0.838}    \\ 
& LS + \alg{} (\textbf{ours}) & 95.34 $\pm$ 0.080  &  \textbf{79.08 $\pm$ 0.307}       & 73.59 $\pm$ 0.623    \\ \bottomrule
\end{tabular}
\end{center}
\caption{Under similar setups, \alg{} consistently achieves higher mean accuracy than Batch Learning across all evaluated benchmarks, a property not shared by other baselines.}
\label{tab:lilac_baselines}
\end{table}
\begin{table}[t!]
\begin{center}
\begin{tabular}{@{}lc@{}}
\toprule
Method & CIFAR-10\\
\midrule
Wide Residual Networks~\cite{zagoruyko2016wide}                  & 96.11  \\
Multilevel Residual Networks~\cite{zhang2017residual}            & 96.23  \\
Fractional Max-pooling~\cite{graham2014fractional}               & 96.53  \\
Densely Connected Convolutional Networks~\cite{huang2017densely} & 96.54  \\
Drop-Activation~\cite{liang2018drop}                             & 96.55  \\
Shake-Drop~\cite{yamada2018shakedrop}                            & 96.59  \\
\textbf{Shake-Drop + \alg{} (ours)}                                & \textbf{96.79}  \\
\bottomrule
\end{tabular}
\end{center}
\caption{\alg{} easily outperforms the Shake-Drop network~(\cite{yamada2018shakedrop}) as well as other top performing algorithms on CIFAR-10 with \textit{standard pre-processing (random crop + flip)}.}
\label{tab:comparison_best}
\end{table}

The RA baseline highlights the importance of using all of the data in $\mathbb{U}$ as negative samples in the IL phase as opposed to using data from individual classes.
This is reflected in the boost in performance offered by \alg.
The DBS baseline is used to highlight the importance of fluctuating mini-batch sizes, which occur due to the balancing of data in the IL phase.
Even with the availability of all labels and fluctuating batch sizes, the DBS baseline is easily outperformed by \alg.
This indicates the importance of the recursive structure used to introduce data in the IL phase as well as the use of data from $\mathbb{U}$ as negative samples.
Overall, \alg{} consistently outperforms Batch Learning across all benchmarks while existing comparable methods fail to do so.
When we extend \alg{} to the Shake-Drop~\cite{yamada2018shakedrop} network architecture, with only standard pre-processing, we easily outperform other existing approaches with comparable setups, as shown in Table~\ref{tab:comparison_best}.

\subsection{Key Properties of \alg}
\label{sec:key_properties_of_lilac}

% \begin{figure}[t!]
%     \centering
%     \includegraphics[width=\columnwidth]{images/prop_epsilon.pdf}
%     \caption{\textcolor{red}{The mid-range of $\epsilon$ values, 0.7-0.4, show a perceptible increase in performance while the edges for either too sharp or too flat a distribution leading to a decrease in performance}}
%     \label{fig:epsilon_trend}
% \end{figure}
\paragraph{Smoothness of Target Vector ($\epsilon$)}
Throughout this work, we maintained the importance of using a smoother distribution as the alternate target vector during the AC phase.
Table~\ref{table:property}~(\textbf{Top}) illustrates the change in performance across varying degrees of smoothness in the alternate target vector.
There is a clear increase in performance when $\epsilon$ values are between 0.7-0.4 (mid-range). 
On either side of this band of values, the GT vector is either too sharp or too flat, leading to a drop in performance.

\paragraph{Size of Label Groups ($m$)}
\alg{} is designed to introduce as many or as few new labels as desired in the IL phase.
We hypothesized that developing stronger representations can be facilitated by introducing a small number of new labels while contrasting it against a large variety of negative samples.
Table~\ref{table:property}~(\textbf{Bottom}) supports our hypothesis by illustrating the decrease in performance with an increase in the number of new labels introduced in each interval of the IL phase.
Thus, we introduce two labels each for CIFAR-10 and STL-10 and only one new label per interval for CIFAR-100 throughout the experiments in Table~\ref{tab:lilac_baselines}.

\begin{table}[t!]
\begin{center}
\begin{tabular}{@{}lccc@{}}
\toprule
\multirow{2}{*}{Property}       &  \multicolumn{3}{c}{Performance (\%)}                           \\
                                                 & CIFAR-10          & CIFAR-100         & STL-10            \\
\midrule
$\epsilon = 0.9$  & 95.30 $\pm$ 0.072 & 78.48 $\pm$ 0.328 & 73.57 $\pm$ 0.980 \\
$\epsilon = 0.8$  & 95.34 $\pm$ 0.141 & 78.52 $\pm$ 0.118 & 73.54 $\pm$ 0.984 \\
\rowcolor{Gray}
$\epsilon = 0.7$  & 95.42 $\pm$ 0.189 & 78.72 $\pm$ 0.356 & 73.59 $\pm$ 0.872 \\
\rowcolor{Gray}
$\epsilon = 0.6$  & 95.36 $\pm$ 0.096 & 78.75 $\pm$ 0.180 & \textbf{73.77 $\pm$ 0.838} \\
\rowcolor{Gray}
$\epsilon = 0.5$  & 95.49 $\pm$ 0.207 & 78.88 $\pm$ 0.227 & 73.61 $\pm$ 0.810 \\
\rowcolor{Gray}
$\epsilon = 0.4$  & \textbf{95.52 $\pm$ 0.072} & \textbf{78.88 $\pm$ 0.201} & 73.54 $\pm$ 0.959 \\
$\epsilon = 0.3$  & 95.31 $\pm$ 0.125 & 78.66 $\pm$ 78.66 & 73.59 $\pm$ 0.955 \\
$\epsilon = 0.2$  & 95.36 $\pm$ 0.095 & 78.47 $\pm$ 0.093 & 73.57 $\pm$ 0.963 \\
\midrule
$m$: 1     & 95.32 $\pm$ 0.156 & \textbf{78.73 $\pm$ 0.139} & 73.27 $\pm$ 0.220 \\
$m$: 2 (4) & \textbf{95.38 $\pm$ 0.135} & 78.34 $\pm$ 0.209 & \textbf{73.43 $\pm$ 0.903} \\ 
$m$: 4 (8) & 95.29 $\pm$ 0.069 & 78.37 $\pm$ 0.114 & 72.30 $\pm$ 0.543 \\ 
\bottomrule
\end{tabular}
\end{center}
\caption{(\textbf{Top}) The mid-range of $\epsilon$ values, 0.7-0.4, show an increase in performance while the edges, due to either too sharp or too flat a distribution, show decreased performance.
(\textbf{Bottom}) \textit{Only IL} model results illustrate the importance of introducing a small number of new labels in each interval of the IL phase. Values in brackets are for CIFAR-100.}
\label{table:property}
\end{table}

\section{Discussion: Impact of Each Phase}
\label{sec:discussion_impact_of_each_phase}

In this section, we take a closer look at the impact of each phase of \alg{} and how they affect the quality of the learned representations.
We extract features from the second to last layer of ResNet18/34 from 3 different baselines (Batch Learning, \alg{}, and \textit{Only IL}) and use these features to train a linear SVM model.

\begin{figure}[t!]
    \begin{center}
    % \centering
    \subfigure{\includegraphics[width=0.45\columnwidth]{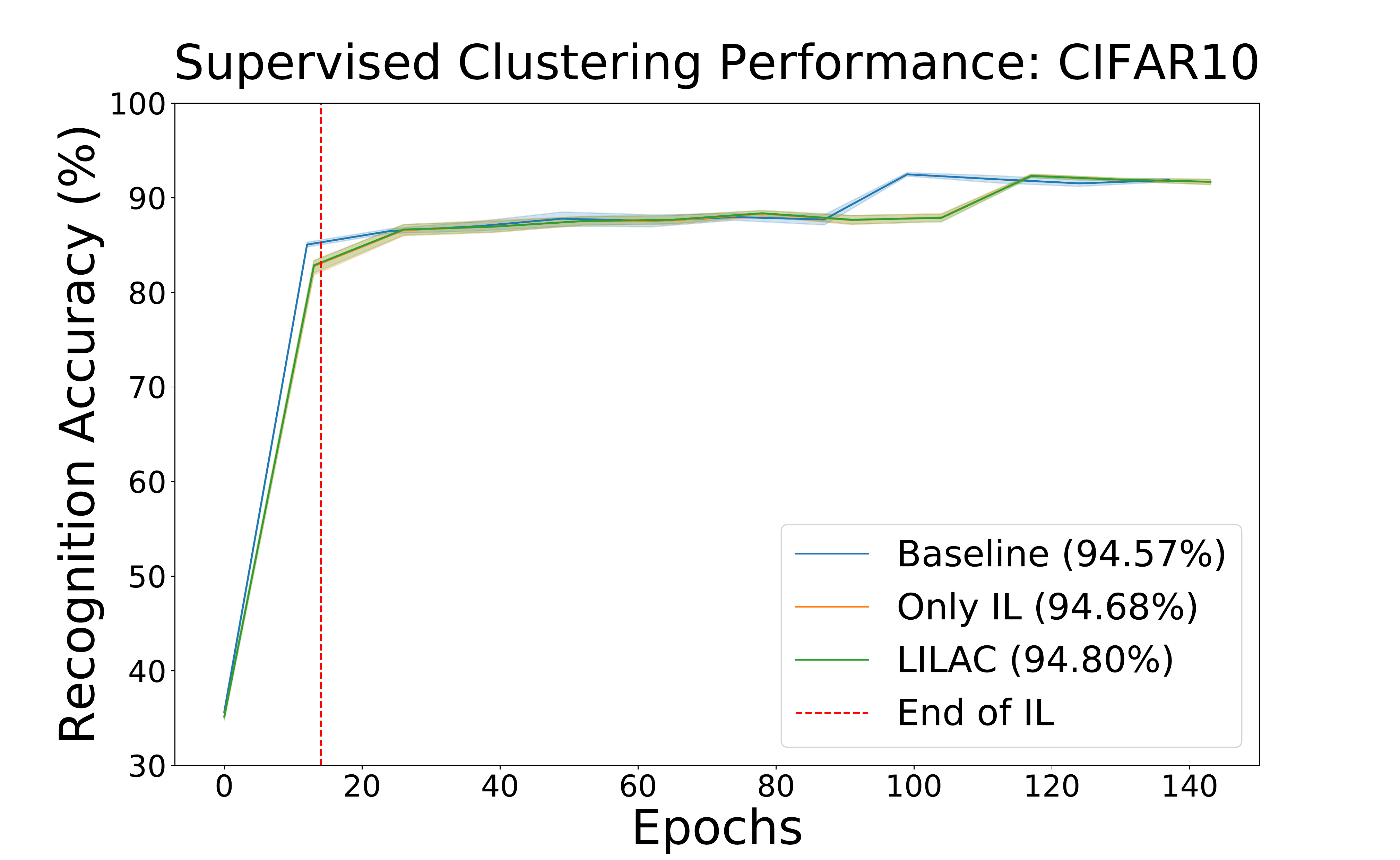}}
    \subfigure{\includegraphics[width=0.45\columnwidth]{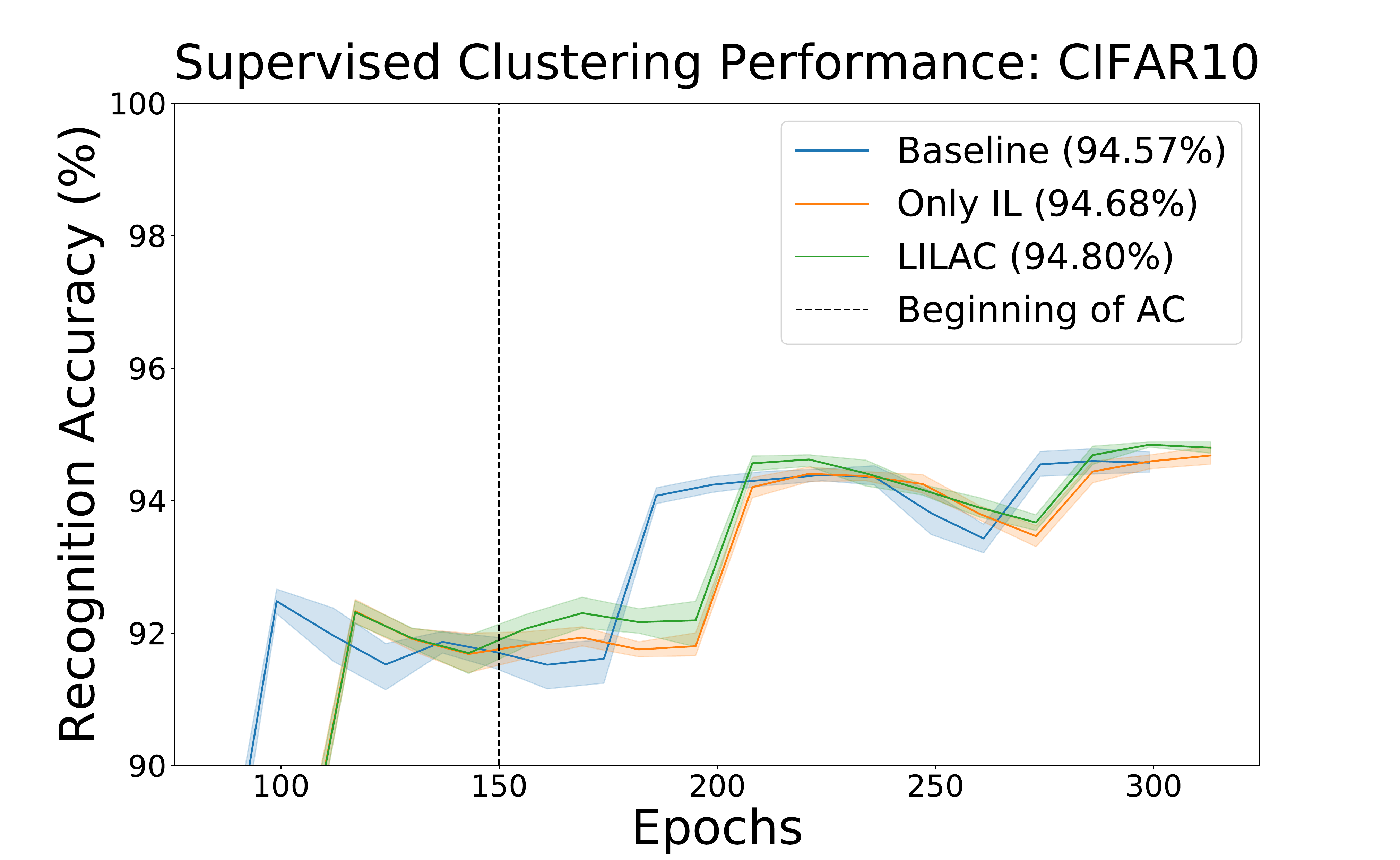}}
    \subfigure{\includegraphics[width=0.43\columnwidth]{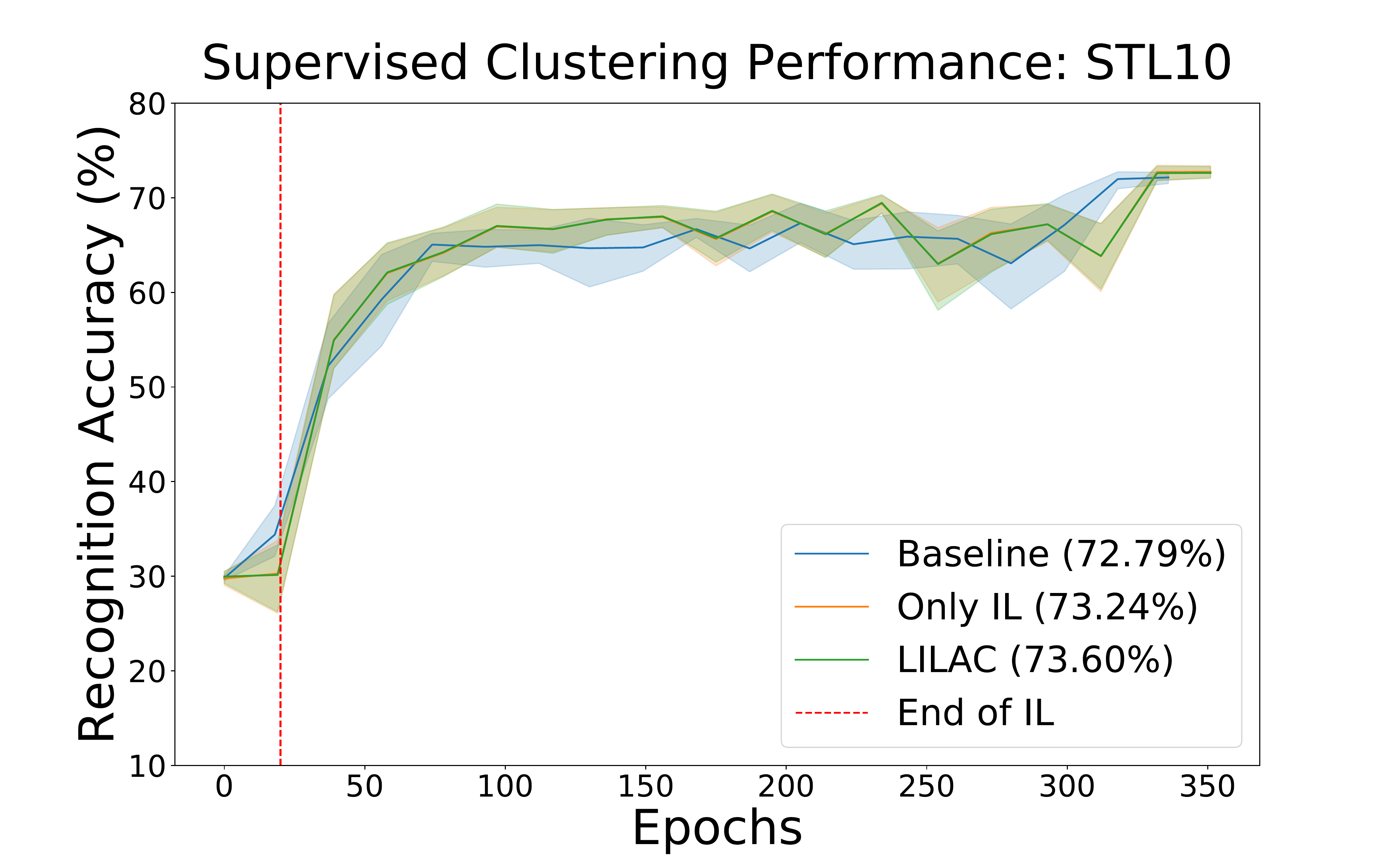}}
    \subfigure{\includegraphics[width=0.43\columnwidth]{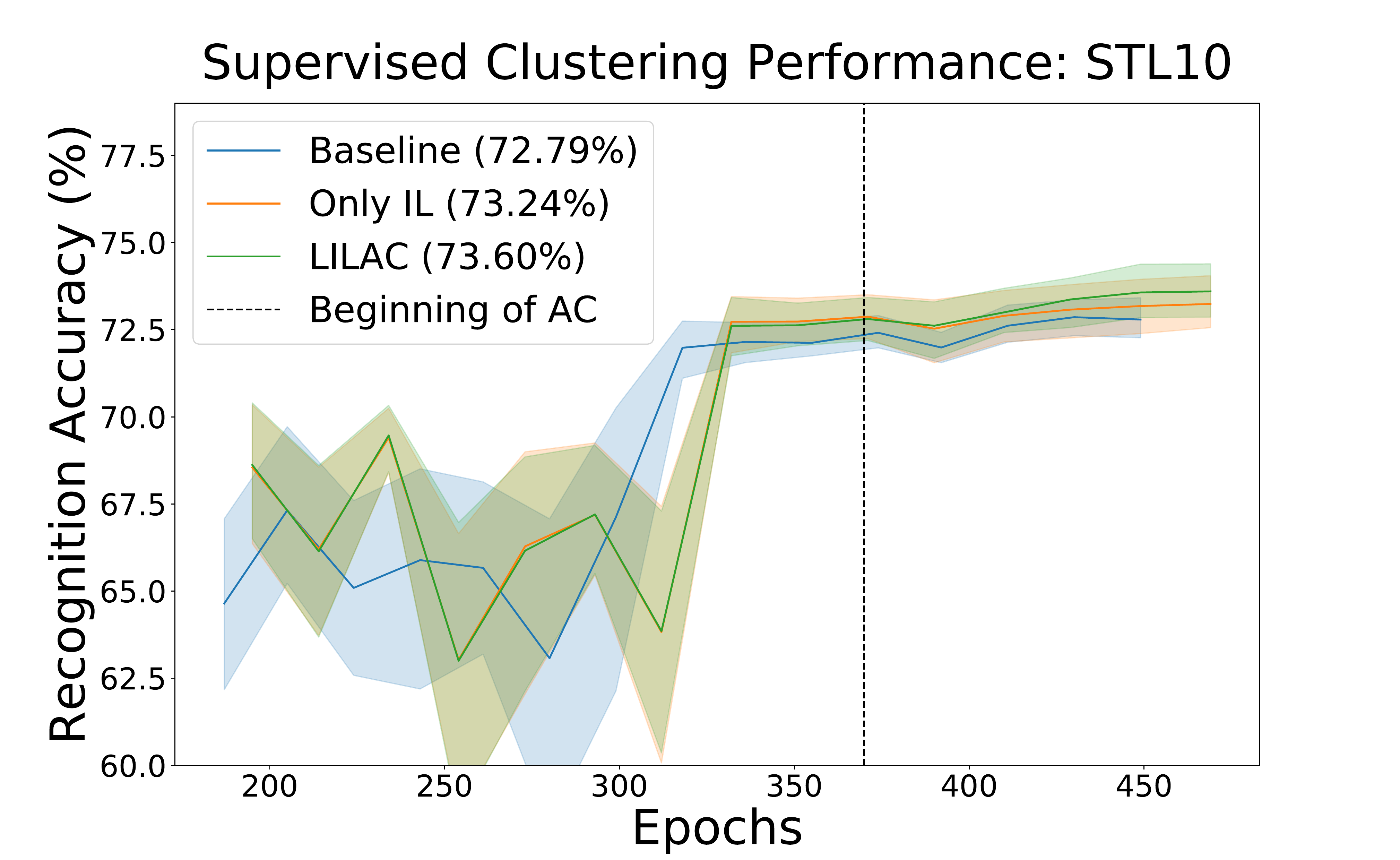}}
    \subfigure{\includegraphics[width=0.45\columnwidth]{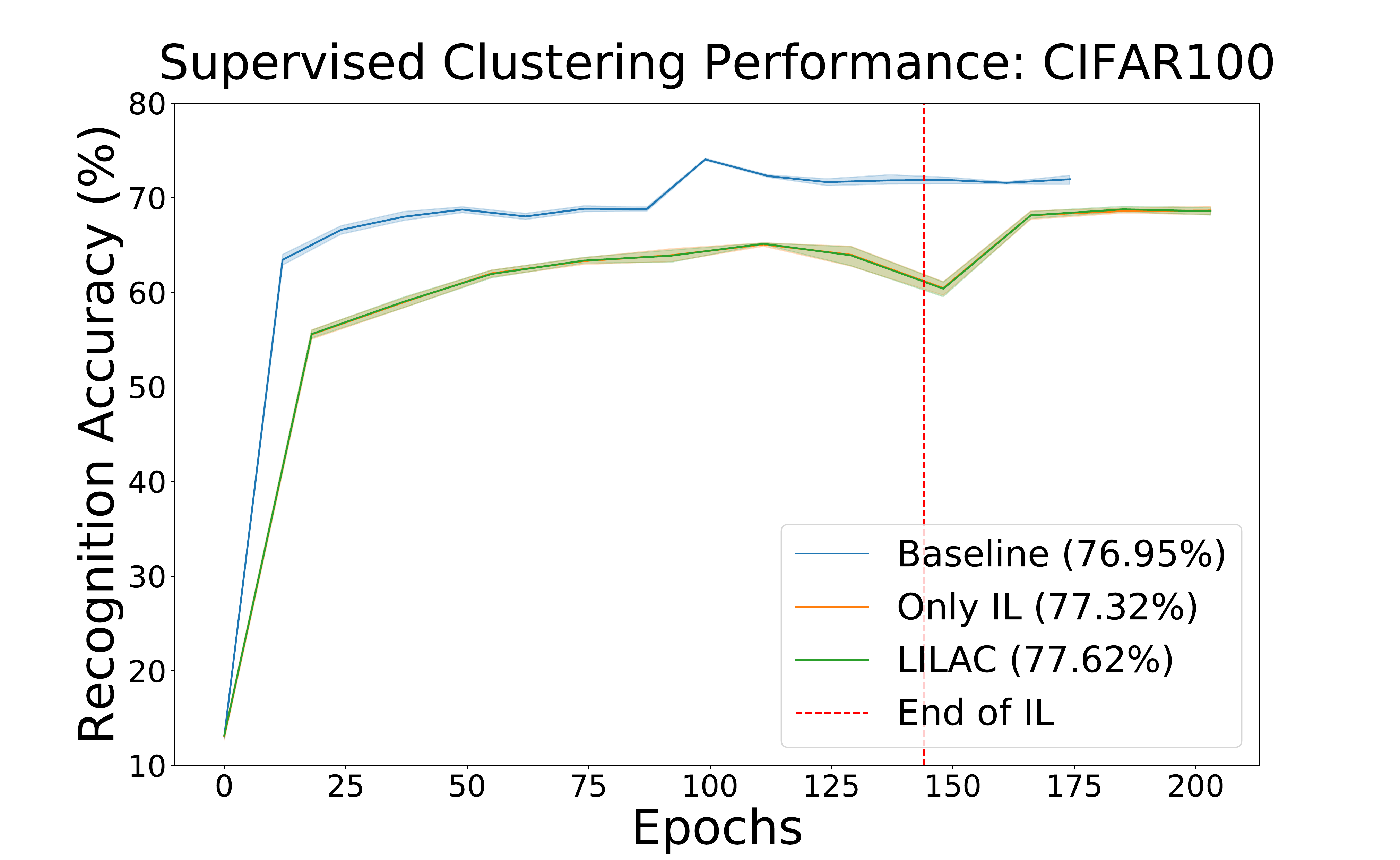}}
    \subfigure{\includegraphics[width=0.45\columnwidth]{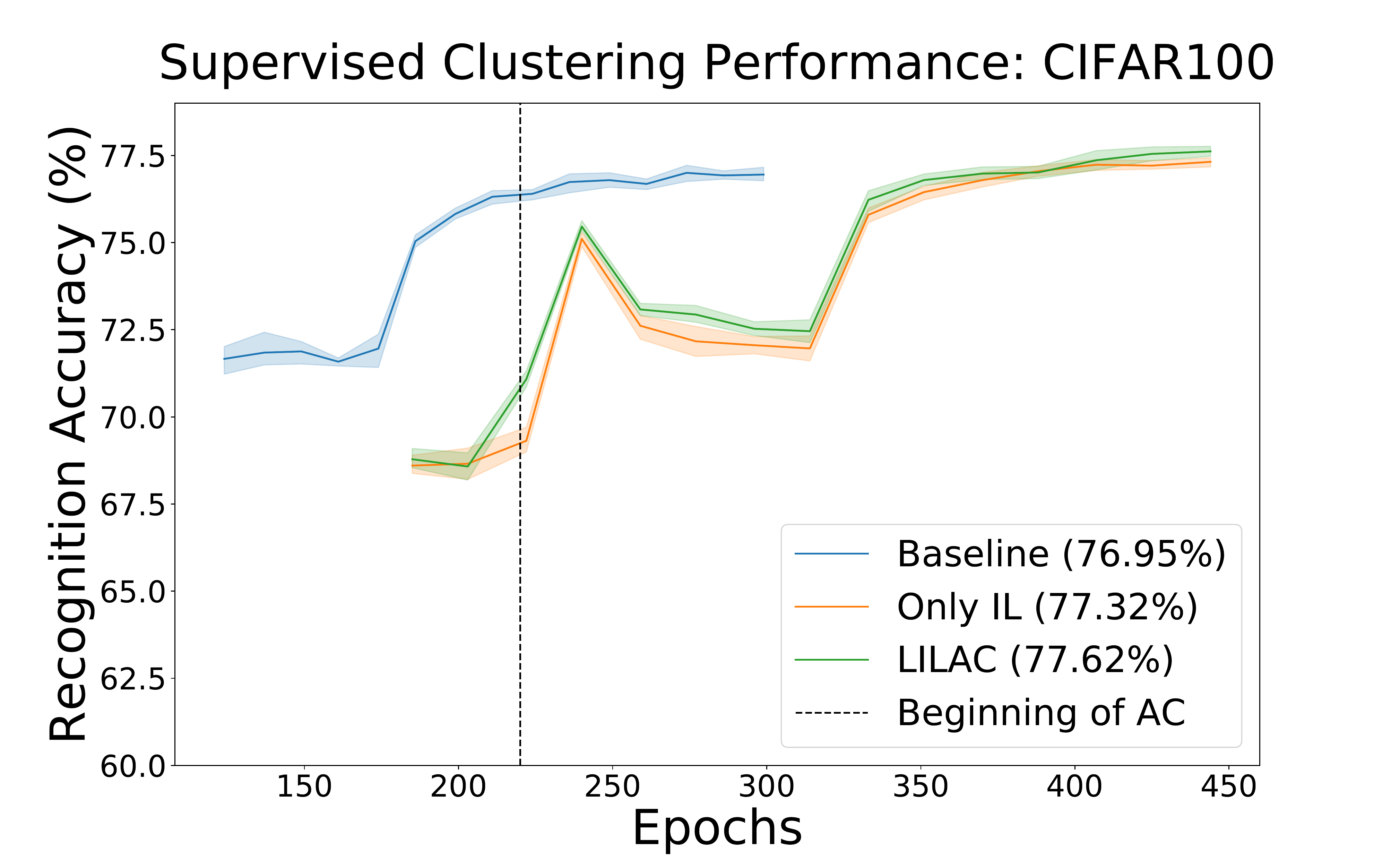}}
    \end{center}
    \caption{Plots on the (\textbf{Left}) show the common learning trend between all baselines, albeit slightly delayed for CIFAR-100, after the IL phase while those on the (\textbf{Right}) show steady improvement in performance after applying AC when compared to the \textit{Only IL} baseline. Final supervised classification performances on representations collected from \alg{} easily outperform those from Batch Learning and \textit{Only IL} methods.}
    \label{fig:representation_comparison}
\end{figure}

Fig.~\ref{fig:representation_comparison} highlights the two important phases in our algorithm.
First, the plots on the left-hand side show a steady improvement in the performance of \alg{} and the \textit{Only IL} baseline once the IL phase is complete and all the labels have been introduced to the network.
When we compare the plots of CIFAR-10 and STL-10 against CIFAR-100, we see that all baselines follow the learning trend shown by Batch Learning, with CIFAR-100 being slightly delayed.
Since there are a large number of epochs required to introduce all the labels of CIFAR-100 to the network, the plots are significantly delayed compared to batch learning.
Conversely, since there are very few epochs in the IL phase of CIFAR-10 and STL-10, we observe the performance trend of \textit{Only IL} and \alg{} quickly match that of Batch Learning.
Overall, the final performances of both \alg{} and the \textit{Only IL} baseline are  higher than Batch Learning, which supports the importance of the IL phase in learning strong representations.

\begin{figure}[ht!]
    \begin{center}
    \subfigure[CIFAR-10]{\includegraphics[width=0.3\textwidth]{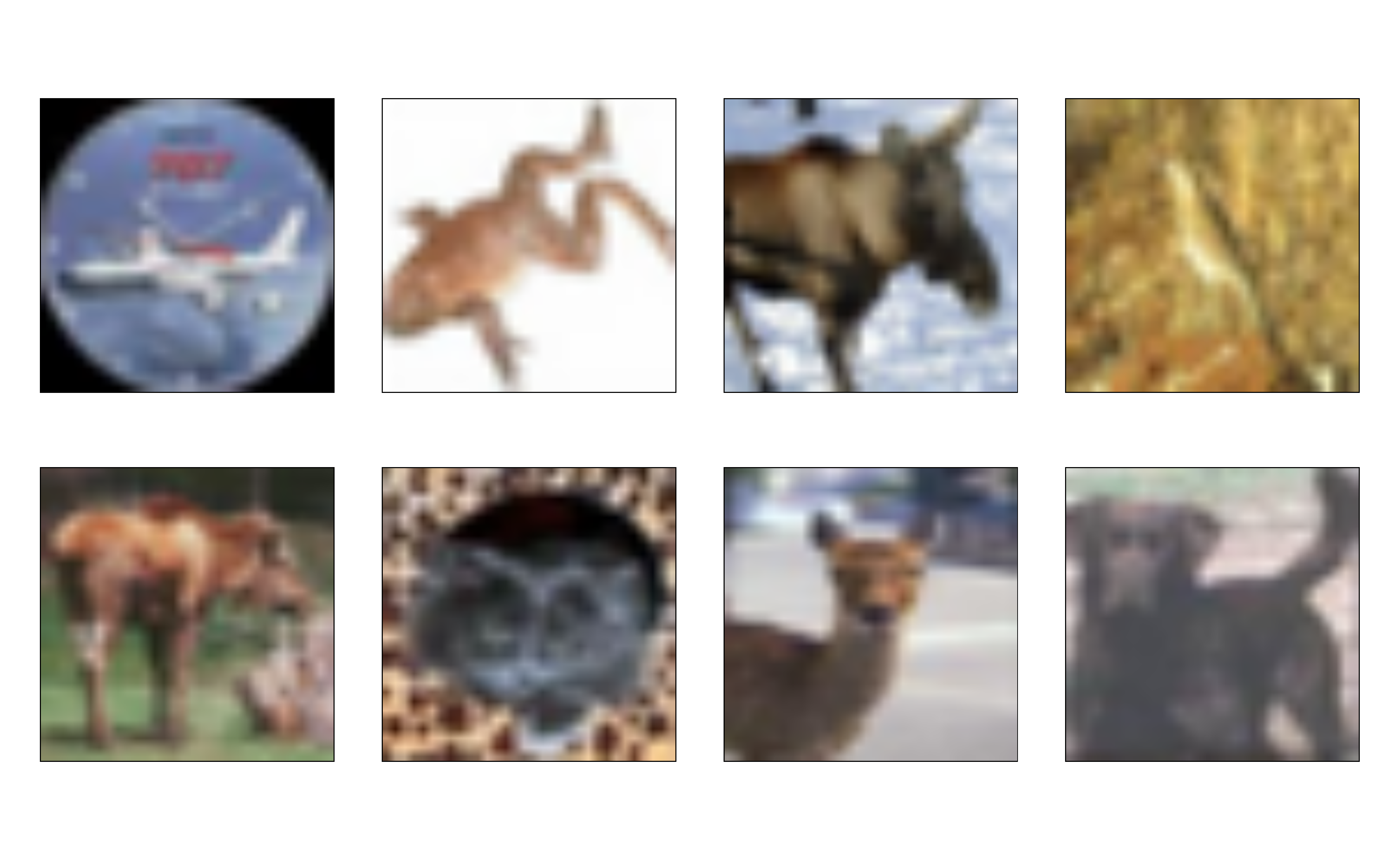}} 
    \subfigure[CIFAR-100]{\includegraphics[width=0.3\textwidth]{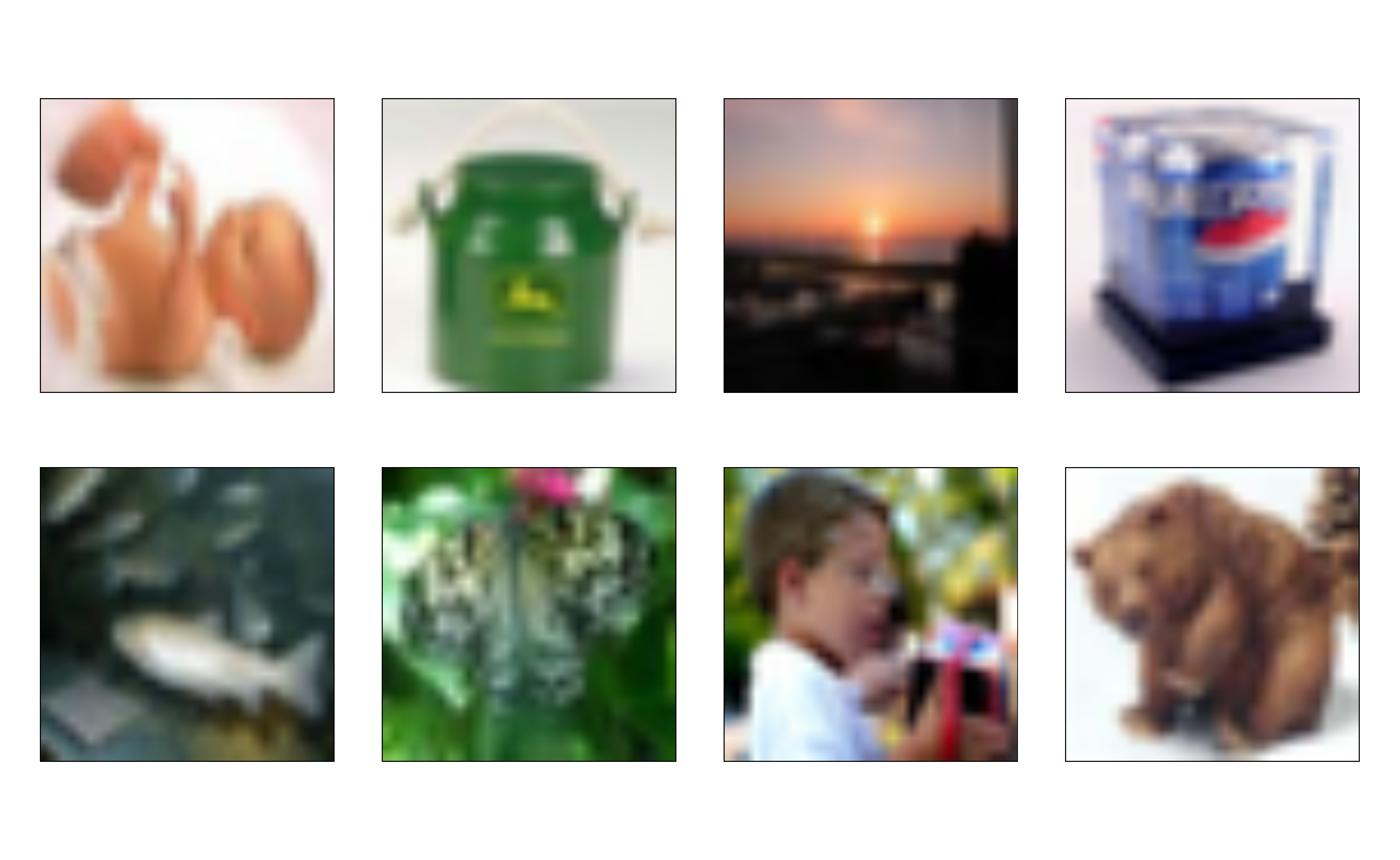}} 
    \subfigure[STL-10]{\includegraphics[width=0.3\textwidth]{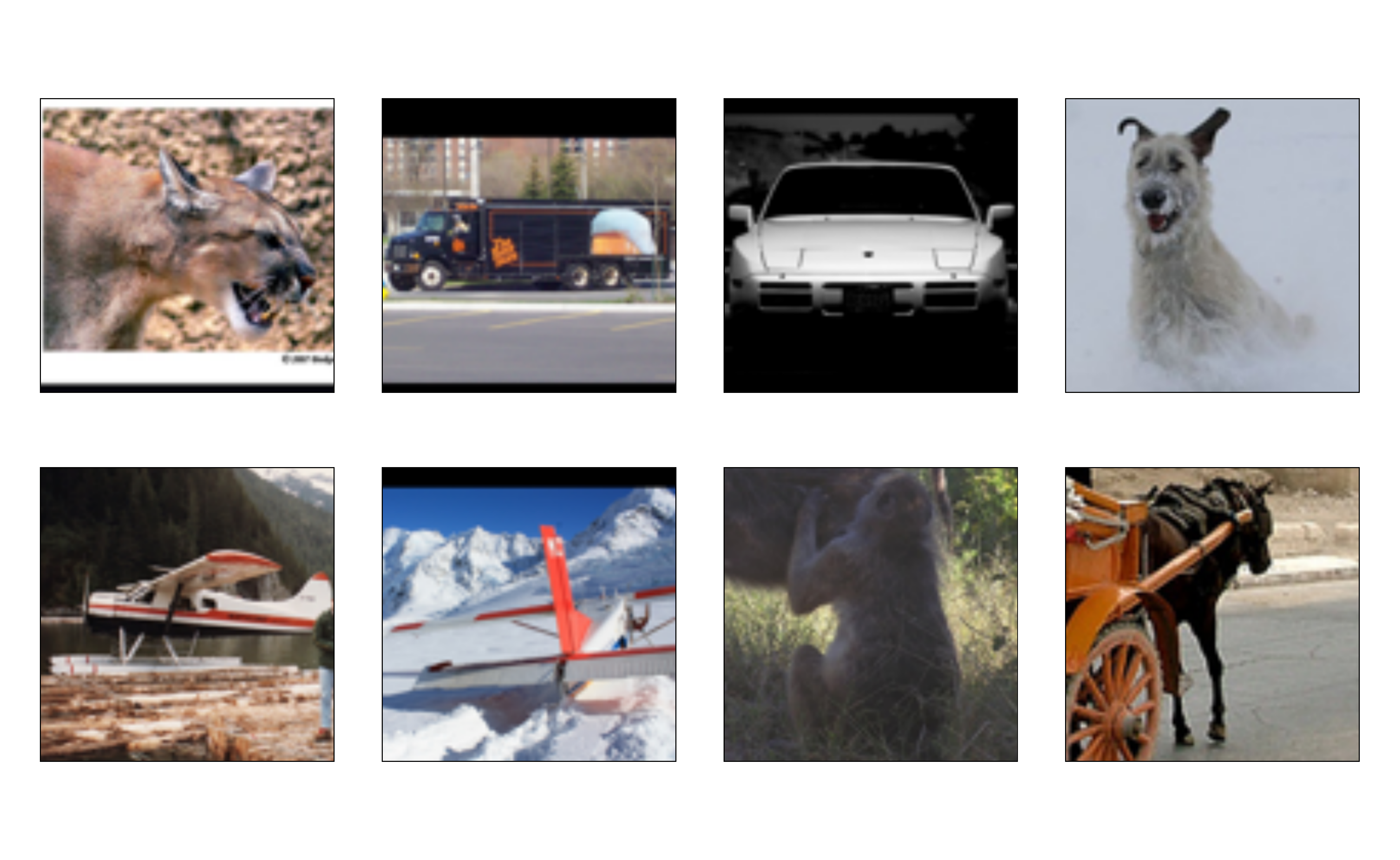}} 
    \end{center}
    \caption{Illustration of 8 randomly chosen samples that were incorrectly labelled by the \textit{Only IL} baseline and correctly labelled by \alg. This highlights the importance of AC.}
    \label{fig:examples_ac}
\end{figure}

\hfill \\
The plots on the right-hand side highlight the similarity in behaviour of \textit{Only IL} and \alg{} before AC.
However, afterward, we observe that the performance of \alg ~overtakes the \textit{Only IL} baseline.
This is a clear indicator of the improvement in representation quality when AC is applied.
Additionally, from Fig.~\ref{fig:representation_comparison} we observe  that inherently the STL-10 dataset results have a high standard deviation, which is reflected in the middle portion of the training phase, between the end of IL and the beginning of AC and it is not a consequence of our approach.
We provide examples in Fig.~\ref{fig:examples_ac} of randomly sampled data from the testing set that were incorrectly classified by the \textit{Only IL} baseline and were correctly classified by \alg.

\section{Conclusion}
\label{sec:conclusion}
In this work, we proposed \alg{}, which rethinks curriculum learning based on incrementally learning labels instead of samples. 
This approach helps kick-start the learning process from a substantially better starting point while making the learned embedding space amenable to adaptive compensation of target vectors.
% Adaptive compensation provides a softer target distribution for misclassified samples which helps correct failed predictions and improve overall performance while decreasing the standard deviation across multiple trials. 
Both these techniques combine well in \alg{} to show the highest performance on CIFAR-10 for simple data augmentations while easily outperforming batch and curriculum learning and label smoothing methods on comparable network architectures.
The next step in unlocking the full potential of this setup is to include a confidence measure on the predictions of the network so that it can handle the effects of dropout or partial inputs.
In further expanding \alg's ability to handle partial inputs, we aim to explore its effect on standard incremental learning (memory-constrained) while also extending its applicability to more complex neural network architectures.

\section{Acknowledgements}
This work was in part supported by NSF NRI IIS 1522904 and NIST 60NANB17D191.  The findings and views represent those of the authors alone and not the funding agencies.
The authors would also like to thank members of the COG lab for their invaluable input in putting together and refining this work.

\bibliography{egbib}

% \documentclass{bmvc2k}

% \title{Supplementary Materials: Rethinking Curriculum Learning with Incremental Labels and Adaptive Compensation}

% % Enter the paper's authors in order
% % \addauthor{Name}{email/homepage}{INSTITUTION_CODE}
% \addauthor{Madan Ravi Ganesh}{http://zeonzir.github.io}{1}
% \addauthor{Jason J. Corso}{https://web.eecs.umich.edu/~jjcorso/}{1}

% % Enter the institutions
% % \addinstitution{Name\\Address}
% \addinstitution{
% University of Michigan\\
% EECS\\
% Ann Arbor\\
% Michigan, USA}

% \runninghead{Ravi Ganesh, Corso}{Label-based Curriculum Learning}

% Any macro definitions you would like to include
% These are not defined in the style file, because they don't begin
% with \bmva, so they might conflict with the user's own macros.
% The \bmvaOneDot macro adds a full stop unless there is one in the
% text already.
% Custom macros
% \newcommand{\cmark}{\ding{51}}
% \newcommand{\alg}{LILAC}
% \newcommand{\xmark}{\ding{55}}
% \DeclareMathOperator*{\argmaxA}{arg\,max} 

% % Custom Packages
% \usepackage{graphicx}
% \usepackage{amsmath,amssymb} % define this before the line numbering.
% \usepackage{multirow}
% \usepackage{color}
% % \usepackage[width=122mm,left=12mm,paperwidth=146mm,height=193mm,top=12mm,paperheight=217mm]{geometry}
% \usepackage{enumitem}
% \usepackage{booktabs}
% \usepackage{bbm}
% \usepackage{pifont}
% \usepackage{bbold}
% \usepackage[ruled,vlined]{algorithm2e}
% \usepackage{subcaption}

\def\eg{\emph{e.g}\bmvaOneDot}
\def\Eg{\emph{E.g}\bmvaOneDot}
\def\etal{\emph{et al}\bmvaOneDot}

%-------------------------------------------------------------------------
% Document starts here
% \begin{document}

% \maketitle

\appendix

\section{Experimental Setup}
In Table~\ref{table:experimental_setup} we list the general hyper-parameters used to train the batch learning portion of every baseline.
This setup covers the training beyond the IL phase for \alg, DBS, RA, and \textit{Only IL} as well as the \textit{Only AC} baseline. 
Across all the methods we ensure that the total number of training epochs, when all the labels in the dataset are known, is held constant.

\begin{table}[h]
\begin{center}
\begin{tabular}{l|l|l}
\hline
Parameters        & CIFAR10/100      & STL10         \\ \hline
Epochs            & 300              & 450           \\ 
Batch Size        & 128              & 128           \\ 
Learning Rate     & 0.1              & 0.1           \\ 
Lr Milestones     & {[}90 180 260{]} & {[}300 400{]} \\ 
Weight Decay      & 0.0005           & 0.0005        \\ 
Nesterov Momentum & Yes              & Yes           \\ 
Gamma             & 0.2              & 0.1           \\ \hline
\end{tabular}
\end{center}
\caption{List of hyper-parameters used to in batch learning. Note: All experiments used the SGD optimizer.}
\label{table:experimental_setup}
\end{table}

\section{Hyper-parameter Selection}
\begin{table}[ht!]
\begin{center}
\begin{tabular}{@{}lccc@{}}
\toprule
\multirow{2}{*}{Property}       &  \multicolumn{3}{c}{Performance (\%)}                           \\
                                                 & CIFAR-10          & CIFAR-100         & STL-10            \\
\midrule
$E = 1 $ & 95.13 $\pm$ 0.175 & 78.21 $\pm$ 0.236 & 72.59 $\pm$ 0.476 \\
$E = 3 $ & 95.20 $\pm$ 0.200 & \textbf{78.73 $\pm$ 0.139} & 73.03 $\pm$ 0.380 \\ 
$E = 5 $ & 95.32 $\pm$ 0.044 & 78.57 $\pm$ 0.102 & 73.08 $\pm$ 0.996 \\ 
$E = 7 $ & \textbf{95.32 $\pm$ 0.156} & 78.44 $\pm$ 0.265 & 73.13 $\pm$ 1.460 \\ 
$E = 10$ & 95.26 $\pm$ 0.185 & 77.98 $\pm$ 0.218 & \textbf{73.27 $\pm$ 0.220} \\ 
\midrule
% \multicolumn{4}{c}{}\\
\midrule
Label Order: Rnd.               & 95.30 $\pm$ 0.146 & 78.35 $\pm$ 0.280 & 73.10 $\pm$ 0.861 \\
Label Order: Difficulty         & 95.25 $\pm$ 0.156 & 78.42 $\pm$ 0.115 & 73.69 $\pm$ 0.849 \\
Label Order: Asc.               & 95.32 $\pm$ 0.156 & 78.73 $\pm$ 0.139 & 73.27 $\pm$ 0.220 \\
\bottomrule
\end{tabular}
\end{center}
\caption{(\textbf{Top}) Varying $E$, the fixed training interval size in the IL phase, shows a dataset specific behaviour, with the dataset with lesser labels preferring a larger number of epochs while the dataset with more labels prefers a smaller number of epochs. 
(\textbf{Bottom}) Comparing random label ordering and difficulty-based label ordering to the ascending order assumption used throughout our experiments, we observe no preference to any ordering pattern.}
\label{table:sm_property}
\end{table}
\begin{figure}[t!]
    \begin{center}
    % \subfigure{\includegraphics[width=0.45\columnwidth]{images/stl10_s_performance_setup1.pdf}}
    % \subfigure{\includegraphics[width=0.45\columnwidth]{images/stl10_s_performance_setup3.pdf}}
    
    \subfigure{\includegraphics[width=0.45\columnwidth]{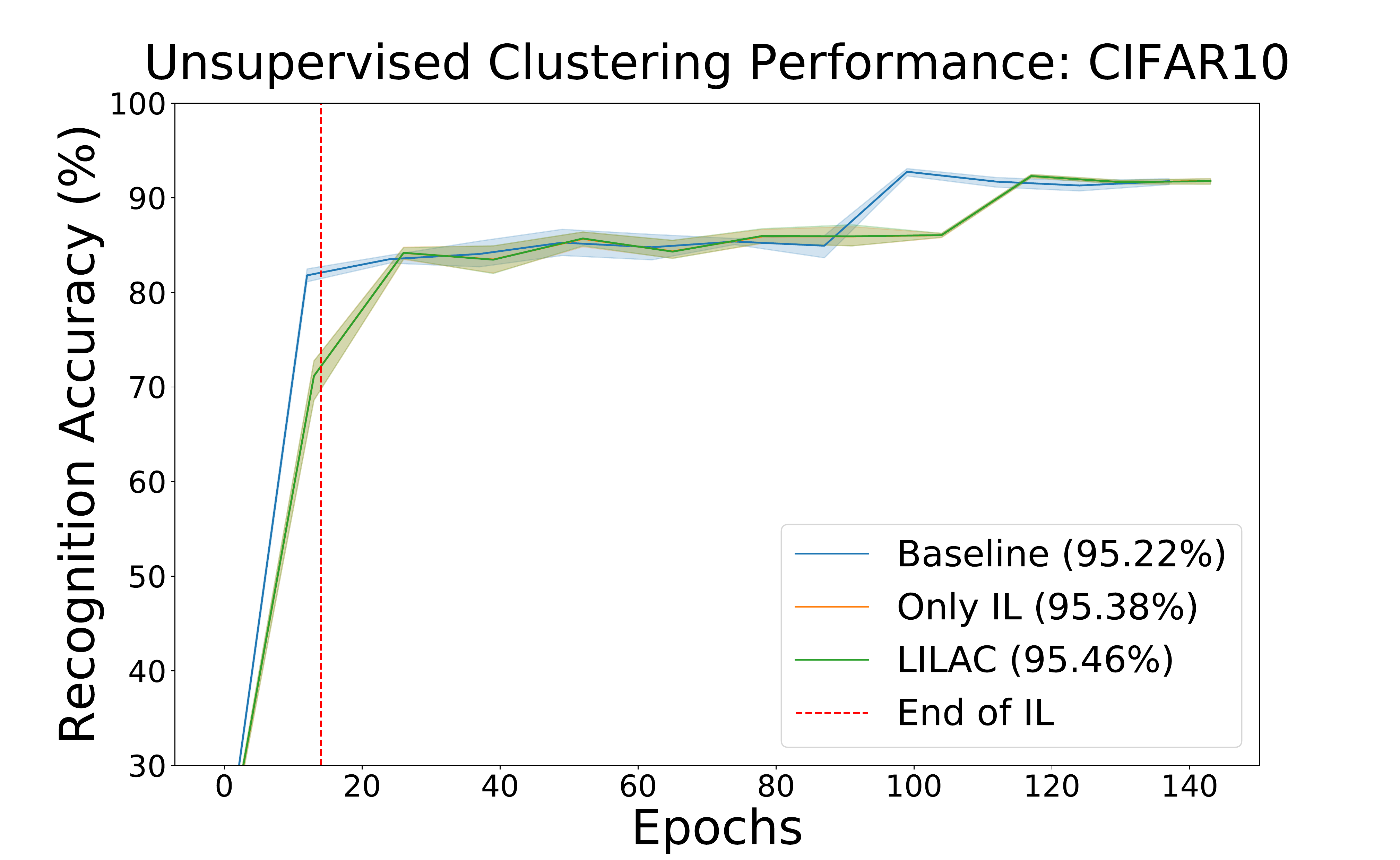}}
    \subfigure{\includegraphics[width=0.45\columnwidth]{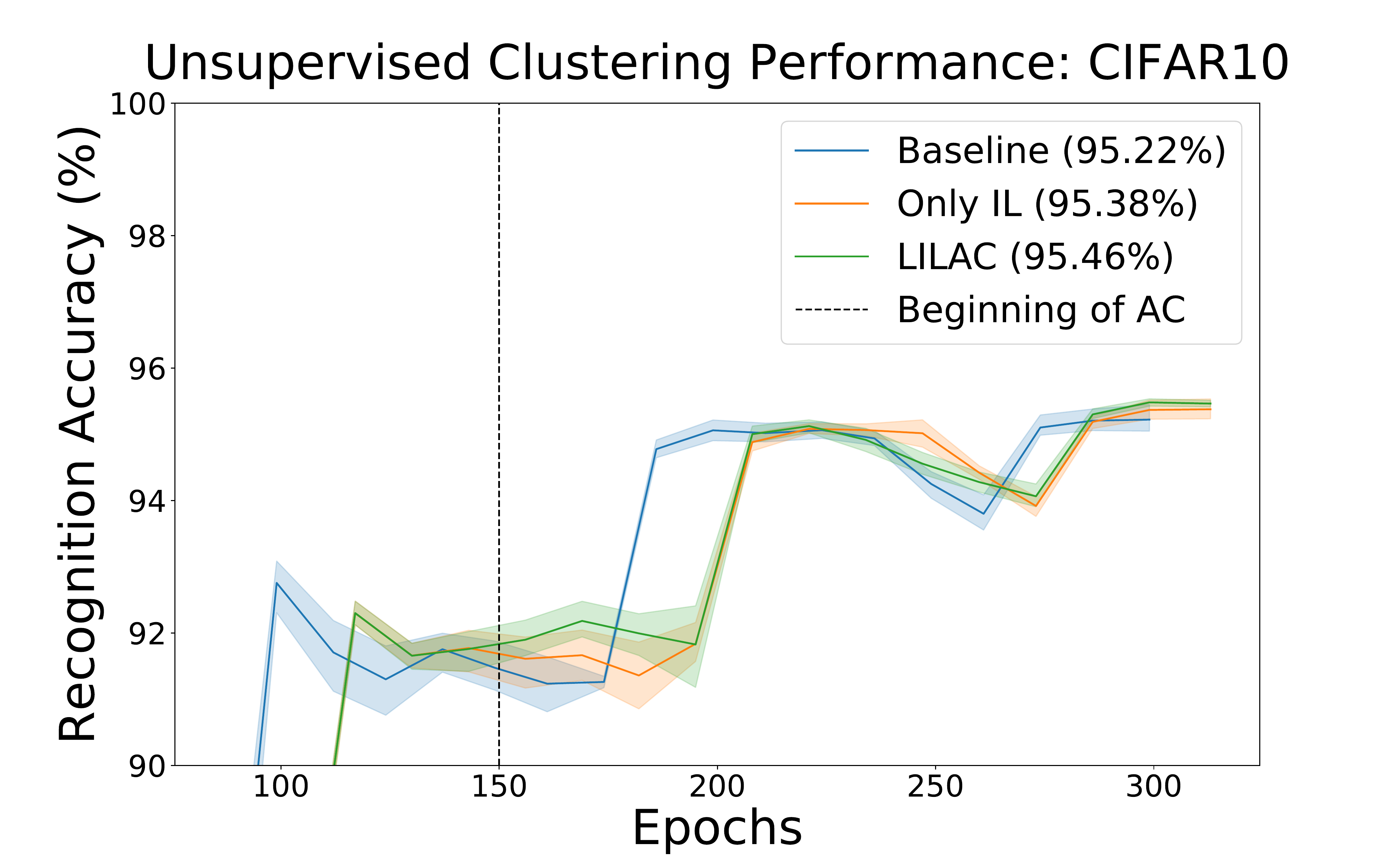}}
    
    \subfigure{\includegraphics[width=0.45\columnwidth]{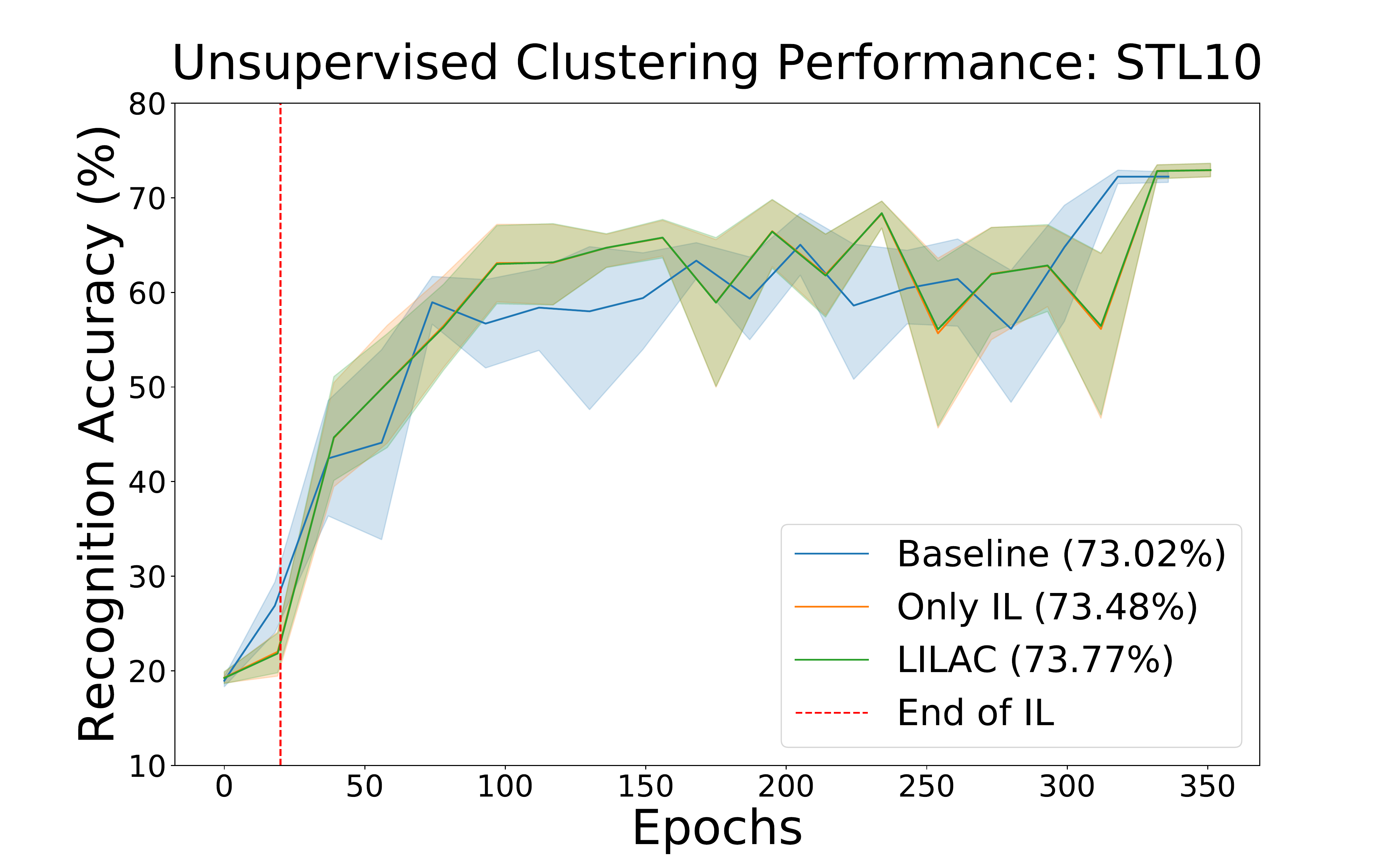}}
    \subfigure{\includegraphics[width=0.45\columnwidth]{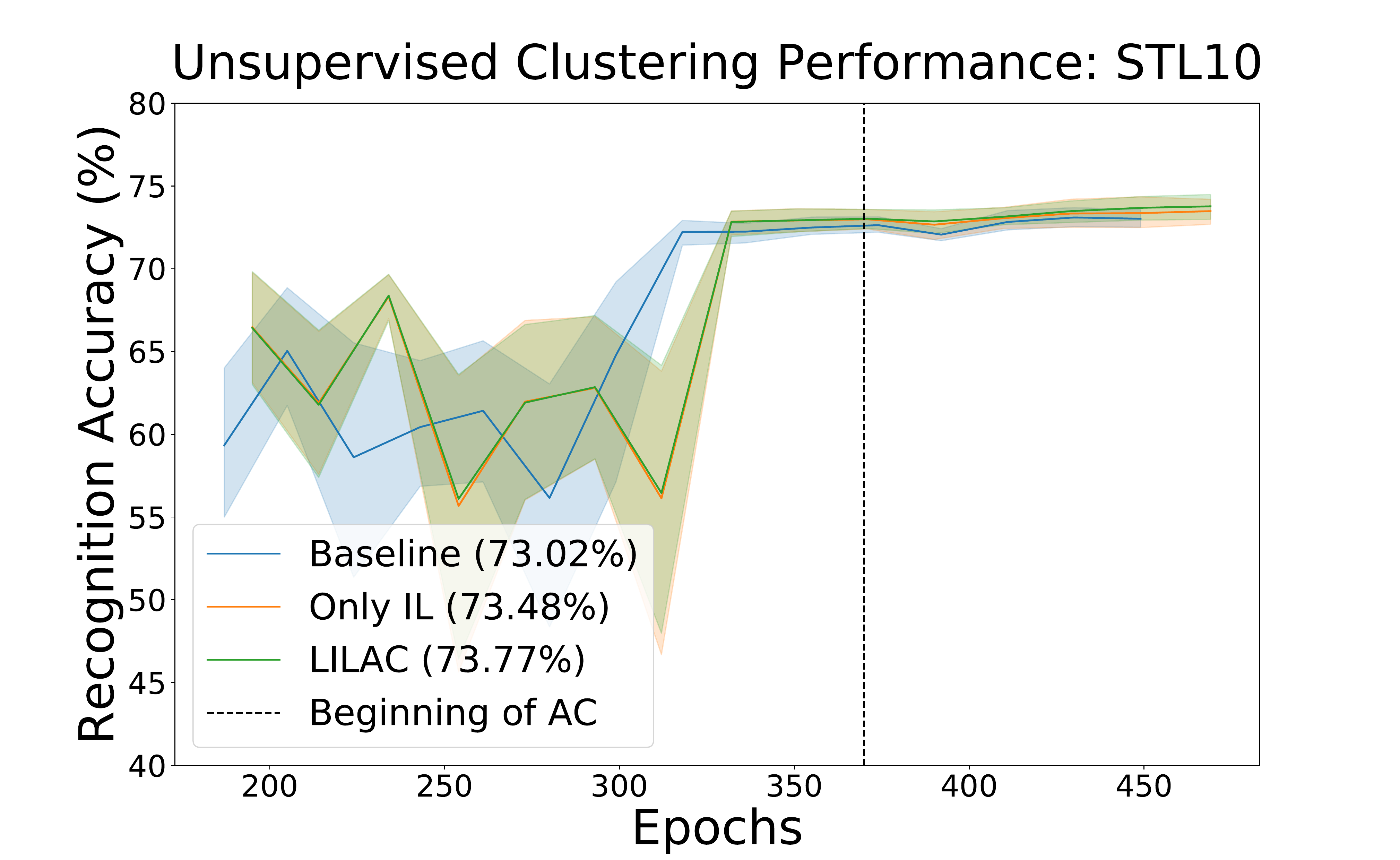}}
    \end{center}
    \caption{Unsupervised classification performance on representations collected from \alg{} easily outperforms those collected from Batch Learning and \textit{Only IL} methods. The plots on the left show the common learning trend between all baselines after IL while plots on the right show steady improvement in performance after applying AC when compared to the baselines.}
    \label{fig:sm_representation_comparison}
\end{figure}
\paragraph{Epochs in Training Interval} When we vary $E$, the fixed training interval size in the IL phase, we observe a dataset specific behaviour.
For datasets with lesser number of total labels, a larger number of epochs provides better performance while for datasets with more labels, a smaller number of epochs yields better performance. 
While the alternate learning rate can have a huge impact on this performance, pacing the introduction of new labels, according to the empirical results, can have a tremendous impact on subsequent hyper-parameters used in \alg.

\paragraph{Label Order} In Table~\ref{table:sm_property}, we compare three different orders of label introduction during the IL phase, 1) random label order, 2) difficulty-based label order, and 3) ascending label order.
Here, difficulty-based label order is obtained from the overall classification scores per label, obtained from the features of a trained model.
Although these three orders do not constitute the exhaustive set of possible label orderings, within these three possibilities there is no definitive order that boosts the performance of \alg{} consistently.
Thus, we employ ascending label order throughout our work.
\hfill \\\hfill \\
\noindent NOTE: \textit{Only IL} baseline is used throughout Table~\ref{table:sm_property}.

\section{Extended Results for Discussion: Impact of Each Phase}
We include unsupervised clustering performance for CIFAR-10 and STL-10 using the k-means and the hungarian job assignment algorithm~\cite{kuhn1955hungarian} in Fig.~\ref{fig:sm_representation_comparison}.
They follow similar patterns to their supervised counterparts.
% \bibliography{egbib}

% \end{document}
\end{document}